\documentclass[twocolumn]{article}
\usepackage{graphicx}
\usepackage[english]{babel} 
\usepackage[hmarginratio=1:1,top=32mm,left=1.2cm,right=1cm,columnsep=0.8cm]{geometry} 
\usepackage[labelfont=bf,up]{caption} 
\usepackage{booktabs} 
\usepackage{fancyhdr} 
\usepackage{titling} 
\usepackage{hyperref} 
\usepackage{url}
\usepackage{amsmath}
\usepackage{amssymb}
\usepackage[ruled]{algorithm2e} 
\usepackage{listings}
\usepackage{xcolor}

\newcommand\pythonstyle{ \lstset{
		language=python,
		frame=top,frame=bottom,
		basicstyle=\scriptsize\ttfamily,
		keywordstyle=\bfseries\ttfamily\color{orange},
		stringstyle=\color{green}\ttfamily,
		commentstyle=\color{red}\ttfamily,
		emph={square}, 
		emphstyle=\color{blue}\texttt,
		emph={[2]root,base},
		emphstyle={[2]\color{yac}\texttt},
		showstringspaces=false,
		flexiblecolumns=false,
		tabsize=2,
		numbers=left,
		numberstyle=\tiny,
		numberblanklines=false,
		stepnumber=1,
		numbersep=10pt,
		xleftmargin=15pt
	}}

\newcommand{ \TensorQuant }{{\it TensorQuant} }
\newcommand{ \Tensorflow }{{\it TensorFlow} }
\newcommand{ \Slim }{{\it Slim} }
\newcommand{ \InceptionI }{Inception\,V1 }
\newcommand{ \InceptionII }{Inception\,V3 }
\newcommand{ \ResnetI }{ResNet\,50 }
\newcommand{ \ResnetII }{ResNet\,152 }

\pretitle{\begin{center}\Large\bfseries} 
    \posttitle{\end{center}} 
\title{Article Title} 
\date{}
\author{%
    \textsc{Dominik Marek Loroch} \\[1ex] 
    \normalsize Fraunhofer ITWM \\ 
    \normalsize \href{mailto:loroch@itwm.fhg.de}{loroch@itwm.fhg.de} 
    \and 
    \textsc{Norbert Wehn} \\[1ex] 
    \normalsize TU Kaiserslautern \\ 
    \normalsize \href{mailto:wehn@eit.uni-kl.de}{wehn@eit.uni-kl.de} 
    \and 
    \textsc{Franz-Josef Pfreundt} \\[1ex] 
    \normalsize Fraunhofer ITWM \\ 
    \normalsize \href{mailto:pfreundt@itwm.fhg.de}{pfreundt@itwm.fhg.de} 
    \and 
    \textsc{Janis Keuper} \\[1ex] 
    \normalsize Fraunhofer ITWM \\ 
    \normalsize \href{mailto:keuper@itwm.fhg.de}{keuper@itwm.fhg.de} 
}
\ifx False
\renewcommand{\maketitlehookd}{%
    \begin{abstract}
        \noindent 
        Recent research implies that training and inference of deep neural networks (DNN) can be computed with
        low precision numerical representations of the training/test data, weights and gradients without a general loss in accuracy.
        The benefit of such compact representations is twofold: they allow a significant reduction of the communication bottleneck in distributed DNN training 
        and faster neural network implementations on hardware accelerators like FPGAs.   
        Several quantization methods have been proposed to map the original 32-bit floating point problem to low-bit representations. While most related publications 
        validate the proposed approach on a single DNN topology, it appears to be evident, that the optimal choice of the quantization method and number of
        coding bits is topology dependent. To this end, there is no general theory available, which would allow users to derive the optimal quantization during the design 
        of a DNN topology.\\
        In this paper, we present a quantization tool box for the \Tensorflow framework. \TensorQuant allows a transparent quantization simulation of existing DNN topologies
        during training and inference. \TensorQuant supports generic quantization methods and allows experimental evaluation of the impact of the quantization on single layers as well as on the full topology.
        In a first series of experiments with \TensorQuant, we show an analysis of fix-point quantizations of popular CNN topologies.
    \end{abstract}
}
\fi

\begin{document}
    
\title{TensorQuant - A Simulation Toolbox for Deep Neural Network Quantization}

\ifx False
\author{Dominik Marek Loroch}
\affiliation{%
  \institution{Fraunhofer ITWM}
}
\email{loroch@itwm.fhg.de}

\author{Franz-Josef Pfreundt}
\affiliation{%
  \institution{Fraunhofer ITWM}
}
\email{pfreundt@itwm.fhg.de}

\author{Norbert Wehn}
\affiliation{%
    \institution{TU Kaiserslautern}
}
\email{wehn@eit.uni-kl.de}

\author{Janis Keuper}
\affiliation{%
  \institution{Fraunhofer ITWM}
}
\email{keuper@itwm.fhg.de}

\renewcommand{\shortauthors}{Loroch et al.}
\renewcommand{\shorttitle}{TensorQuant Toolbox}
\fi

\maketitle

\ifx\a\a
\begin{abstract}
Recent research implies that training and inference of deep neural networks (DNN) can be computed with
low precision numerical representations of the training/test data, weights and gradients without a general loss in accuracy.
The benefit of such compact representations is twofold: they allow a significant reduction of the communication bottleneck in distributed DNN training 
and faster neural network implementations on hardware accelerators like FPGAs.   
Several quantization methods have been proposed to map the original 32-bit floating point problem to low-bit representations. While most related publications 
validate the proposed approach on a single DNN topology, it appears to be evident, that the optimal choice of the quantization method and number of
coding bits is topology dependent. To this end, there is no general theory available, which would allow users to derive the optimal quantization during the design 
of a DNN topology.\\
In this paper, we present a quantization tool box for the \Tensorflow framework. \TensorQuant allows a transparent quantization simulation of existing DNN topologies
during training and inference. \TensorQuant supports generic quantization methods and allows experimental evaluation of the impact of the quantization on single layers as well as on the full topology.
In a first series of experiments with \TensorQuant, we show an analysis of fix-point quantizations of popular CNN topologies.              
\end{abstract}
\fi
%
%


\section{Introduction}
\begin{figure} [ht]
    \includegraphics[width=0.45\textwidth]{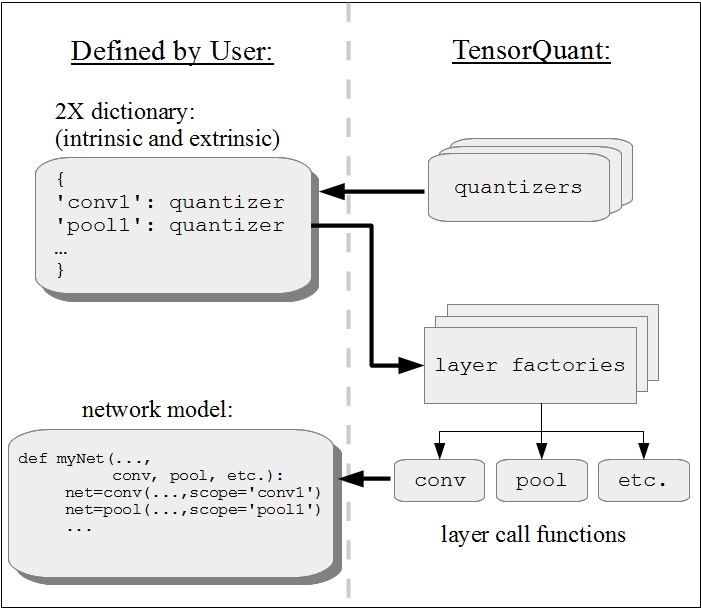}
    \caption{Overview of the \TensorQuant toolbox.}
    \label{fig:TensorQuant}
\end{figure}
Deep Neural Networks suffer from the big amount of data which needs to be stored or transferred during training and inference. The data is usually represented by floating point numbers, because they are the most convenient to handle on standard hardware. However, the required memory, energy and achieved throughput of hardware depend approximately linearly on the number of bits necessary to represent the data. Several publications suggest that the floating point representation is taking more resources than it would be necessary to successfully train networks and perform inference \cite{HanMao15, RasOrd16}.\\
There are two important use cases where smaller data representations are particularly interesting: The first one is custom hardware such as FPGAs \cite{HanLiu16} and ASICs \cite{JouYou17}, where not only data storage and transfer can be carried out in a customized format, but also the computation. The second case is distributed systems \cite{KeuperDL, GupAgr15}, where the communication between the different nodes is becoming the main bottleneck. A custom data representation can largely reduce the amount of data which needs to be communicated and thus reduce energy consumption and increase throughput.\\
A common approach to reduce the amount of data is quantization, which is a mapping from a large, continuous or discrete set of values to a discrete, smaller set.

\subsection{Related Work}
There are several quantization methods which can reduce the amount of stored and transferred data in neural networks. One common approach is to quantize the data with clustering, which means confining the data representation to a discrete set of values \cite{HanMao15, ZhoYao17, GonLiu14}. 
A special case of quantization is the fixed point representation, where all numbers of the discrete set have the same distance to their two nearest neighbours \cite{LiDe17, LinTal16, GupAgr15, JudAlb15}. Fixed point numbers are a very popular representation in custom hardware like \mbox{FPGAs} and \mbox{ASICs}.
An extreme case of quantization are binary \cite{CouHub16, RasOrd16} and ternary quantization \cite{LiZha16, ZhuHan16}. Only one or two bits, respectively, are used to represent a value. Although the compression for the parameters is very high, these methods need gradients represented in floating point while training.
There are various other methods like using the hashing trick \cite{CheWil15}, logarithmic quantization \cite{MiyLee16}, etc.\\
Among these methods, the fixed point representation has gained attention because of the emergence of powerful custom hardware in datacenters, like Google's Tensor Processing Unit \cite{JouYou17}, Amazon Web Services' F1 instances and Microsoft Azure's FPGA-based cloud services.\\
Many publications claim that the rounding, which inevitably happens after every operation in custom hardware using the fixed point format, can be modelled with a single rounding step applied after a DNN layer has been computed in floating point precision \cite{JudAlb15, GupAgr15, LinTal16}. 
The argument in \cite{LinTal16} is that the quantization noise introduced after every rounding step can be transformed into a single source of noise at the end of the layer, since all operations are linear, if the activation function is a rectifying linear unit (ReLU). In other words, it does not matter at which point the noise level is increased, thus rounding at the end is a sufficient approximation.\\

\subsection{Contribution}
In this paper, the question whether or not the quantization noise introduced after each operation is close to a single quantization step at the end of a layer is answered by directly simulating those two cases and comparing them with each other.
In addition to the location of the rounding, the rounding method is also investigated. 
Even simple rounding methods, such as nearest rounding and rounding down (defined in section \ref{sec:fixed_point}), have an impact on the network's accuracy.\\
Quantization methods are often tested on simple datasets and small topologies, which can be quickly trained and run, such as LeNet and AlexNet with MNIST or CIFAR10. The results from these experiments are generalized to bigger networks and datasets. Here, the results are directly obtained from simulating big, state-of-the-art topologies, such as \InceptionII and \ResnetII.\\
In order to investigate quantization in DNNs, a toolbox for \Tensorflow called \TensorQuant is introduced\footnote{available at: \url{https://github.com/cc-hpc-itwm/TensorQuant}}. The full spectrum of functions offered by \Tensorflow can be utilized, augmented with the ability to quantize each layer and to fully emulate fixed point format data processing.
Up until now, there was no implementation which could emulate custom size fixed point computation in a common neural network simulation framework.
In short, the main contributions are:
\begin{itemize}
	\item A toolbox for \Tensorflow, which can quantize the user's network using any user defined quantization method.
	\item Emulation of fixed point operations in addition to layerwise quantization.
    \item A systematic investigation of the impact of the rounding location and method (the ''where'' and ''how'') on the DNN accuracy.
	\item A demonstration of the capabilities of the toolbox on large, state-of-the-art networks.
\end{itemize}
This paper is structured as follows: Chapter 2 introduces the used methods and terms. Extrinsic and intrinsic rounding are introduced and explained, which have a central role. Chapter 3 presents the \TensorQuant toolbox and explains its features in detail. It gives an overview on how much effort the user needs to put into applying the toolbox to his or her own projects. The toolbox is used to investigate fixed point quantization in chapter 4. Experiments are carried out on large topologies such as \InceptionII and \ResnetII. 

\section{Terms and Methods}
\label{sec:Methods}

\subsection{Fixed Point Representation and Rounding:}
\label{sec:fixed_point}
A fixed point number $(W,F)$ is an integer with word size $W$, where the $F$ least significant bits are interpreted as the fractional portion of the number. The word size or width $W$ is defined as the number of bits which are used to store a single numerical value. Negative numbers are saved in two's complement, thus the range of a fixed point number $(W,F)$ is
\begin{equation} \label{eq:fp_range}
[-2^{W-F-1},\,2^{W-F-1}-2^{-F}].
\end{equation}
The resolution $\Delta$ of the fixed point number is $2^{-F}$.\\
Converting a number from floating point representation to fixed point causes loss in accuracy. If the original number does not fit into the fixed point range determined by equation \ref{eq:fp_range}, the usual approach is to saturate the number, that is to use the positive or negative marginal value, respectively.\\
The fractional part is chosen using a rounding method. Commonly known is nearest rounding
\begin{equation} \label{eq:rounding_nearest}
Q_{\text{nearest}}(x)=\text{sgn}(x)\lfloor \frac{\lvert x \rvert}{\Delta}+\text{0.5}\rfloor\Delta,
\end{equation}
where the number $x$ is rounded to the closest element in the finite set of fixed point numbers with the resolution of $\Delta$.\\
Another way is to round towards zero, that is to throw away the fractional part of the number after the given resolution:
\begin{equation} \label{eq:rounding_zero}
Q_{\text{zero}}(x)=\text{sgn}(x)\lfloor \frac{\lvert x \rvert}{\Delta}\rfloor\Delta.
\end{equation}
The easiest way to implement rounding in custom hardware is down rounding
\begin{equation} \label{eq:rounding_down}
Q_{\text{down}}(x)=\lfloor \frac{ x }{\Delta}\rfloor\Delta,
\end{equation}
which cuts off the binary representation of the number after as many bits as corresponds to $\Delta$, regardless of the sign. For example, if $\Delta=$0.25, the binary number is cut off after the second fractional bit.\\
A rounding method which has been investigated often in literature is stochastic rounding \cite{GupAgr15,LiDe17}
\begin{equation} \label{eq:rounding_stochastic}
Q_{\text{stochastic}}(x)= \left\lbrace \begin{array}{ll}
\lceil\frac{ x }{\Delta}\rceil \Delta & \text{if}\quad  \frac{x - \lfloor x \rfloor}{\Delta} >= t_\text{random} \\
\\
\lfloor\frac{ x }{\Delta}\rfloor \Delta & \text{otherwise}
\end{array}
\right. ,
\end{equation}
where the number $x$ is rounded up or down depending on the random, uniformly distributed threshold $t_\text{random}\in \left[0,1\right]$. The probability to be rounded towards either of the neighbouring values increases with the proximity to that value.

\subsection{Extrinsic, Intrinsic and Gradient Quantization:} 
\begin{figure} [ht]
    \includegraphics[width=0.45\textwidth]{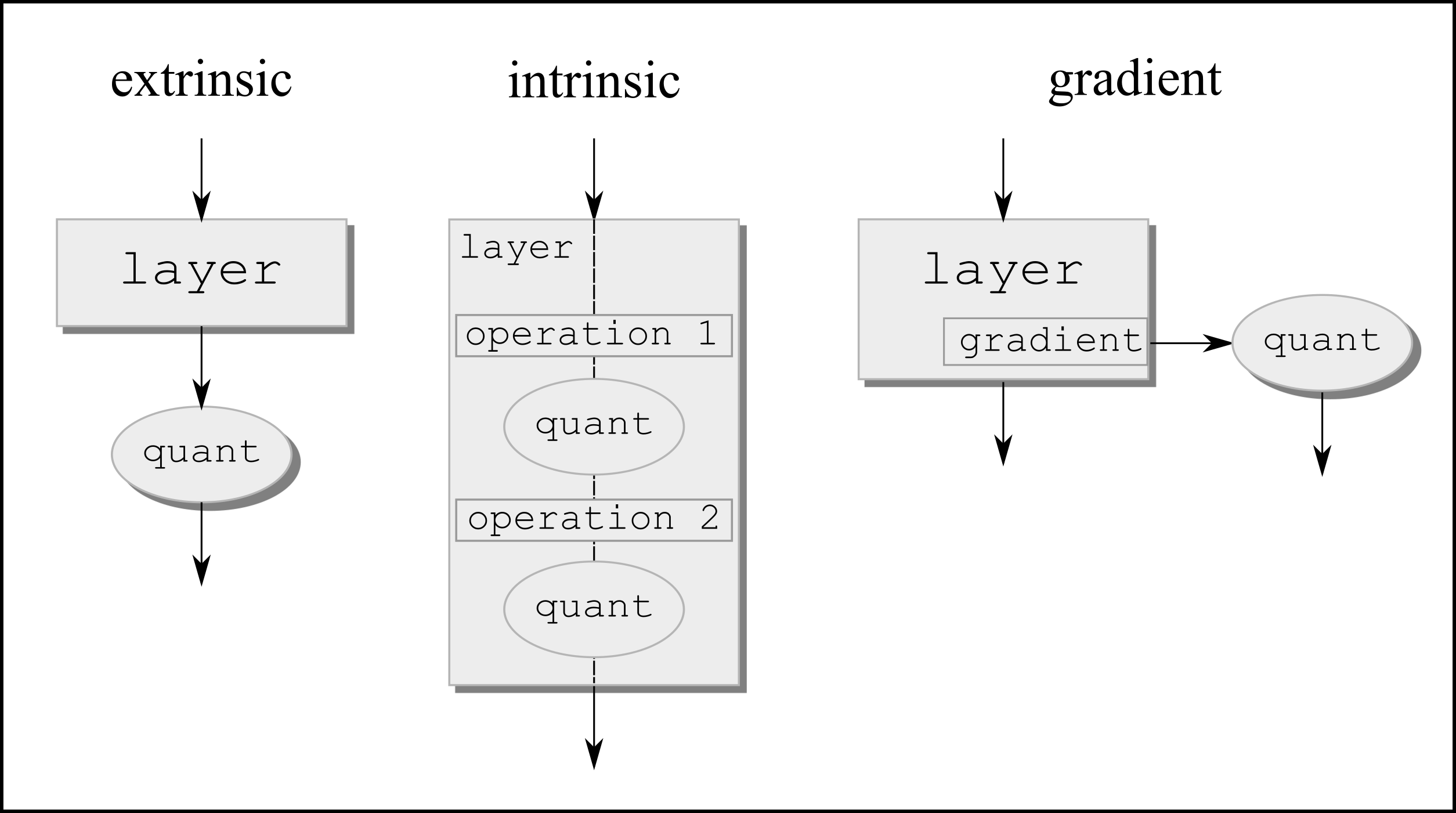}
    \caption{Overview of quantization types.}
    \label{fig:quantization_types}
\end{figure}
In order to handle the question of where the quantization is applied to, as it was described in the introduction, the terms ''extrinsic'' and ''intrinsic'' quantization are introduced in this paper (see figure \ref{fig:quantization_types} for an overview).\\
If quantization is applied at the end of a layer, it will be called extrinsic quantization. This is the case where a layer is computed in floating point precision, but the output is sent to the next layer in a reduced data format. 
An extrinsic quantization scenario can be encountered in a distributed system with many computation nodes, where the data is processed in floating-point precision within the nodes, but it is quantized before it is sent to another node in order to reduce the required bandwidth.
For example, a convolution layer, which performs the operation $\circledast$ between the input $x_\text{in}$ and the filter $K$, would be formally represented as
\begin{equation} \label{eq:extr_quantization}
x_\text{in} \circledast K= Q( \sum_{f}\sum_{k}\sum_{l}x_{\text{in},f}(i+k,j+l)\times K_{f}(k,l) ),
\end{equation}
with $Q$ the quantization function, $f$ the number of features, $k$ and $l$ the coordinates within the filter and $i$ and $j$ the coordinates within a feature.\\
On the other hand, all intermediate results from every arithmetic operation within a layer can be quantized. This scenario can be found in custom hardware, where data is stored and processed in the fixed point format, thus the data is restricted to a certain precision at any point. 
Since this type of quantization is applied on a deeper level compared to the extrinsic one, it is called intrinsic quantization.
A convolution layer for example is formally computed as
\begin{equation} \label{eq:intr_quantization}
x_\text{in} \circledast K= Q( \sum_{f}\sum_{k}\sum_{l} Q(Q(x_{\text{in},f}(i+k,j+l))\times Q(K_{f}(k,l))) ).
\end{equation}
Simulating intrinsic quantization needs more memory and time then the extrinsic case. It is only reasonable to use intrinsic quantization when trying to emulate the behaviour of hardware. Fixed point quantization is the most reasonable method to apply intrinsically, which is simply referred to as rounding in this paper.\\
During training, the gradient can be quantized before the update to the weights is applied \cite{KeuperDL}. 
Formally, this is can be described with
\begin{equation} \label{eq:grad_quantization}
w_\text{t+1}=w_\text{t}+\eta Q(\nabla L),
\end{equation}
with $w$ the trained parameter at a time $\text{t}$, $\eta$ the learning rate and $\nabla L$ the gradient of the loss function, which is to be minimized.\\

\subsection{The Subunit Quantization Approach:} 
\label{sec:subunit_quantization}
Intrinsic quantization can be unfeasible for big DNN topologies such as \InceptionII \cite{SzeVan16} and \ResnetII \cite{HeZha16}, because it requires too much memory to run. A method to mitigate this problem is to apply rounding only to functional subunits of the topology. A good example for subunits are the inception modules occurring in the Inception type topologies. 
So instead of rounding all layers at once, only a few are rounded at the same time. There are good reasons justifying this approach. First, the accuracy loss which happens at one quantized layer cannot be regained in subsequent layers. Second, it is very likely that the various layers require different word sizes and fractional bits in order to keep the accuracy at the baseline value. Therefore, there exists one or several bottlenecks, which will determine the word size of the entire topology.\\
All subunits of the topology are rounded one after another and with different word sizes and fractional bits. The accuracy of the inference is recorded for each run. For each subunit, the best combination with the lowest word size and the least fractional bits is determined, for which the accuracy stays the same compared to the unquantized topology. Amongst all the best combinations, the subunit with the highest word size is identified as the bottleneck. If there are several bottleneck subunits with the same word size, the unit with the highest fractional bits is chosen as the bottleneck.\\

\section{The TensorQuant Toolbox}
The \TensorQuant toolbox is able to quantize any neural network designed in \Tensorflow intrinsically and extrinsically. Some changes to the user's original \Tensorflow topology description file need to be made. Also, the user has to provide a specification of which layers shall be quantized. \TensorQuant manages the quantization of the layers during the building and running process. An overview of the different components of \TensorQuant is given in figure \ref{fig:TensorQuant}.

\paragraph{The Quantizers}
\pythonstyle
\begin{figure}
    \begin{lstlisting}[caption={Fixed point quantizer code.},label={lst:quantizer}]
class FixedPointQuantizer(Quantizer_if):
    def __init__(self, fixed_size, fixed_prec):
        self.fixed_size=fixed_size
        self.fixed_prec=fixed_prec
    def quantize(self,tensor):
        return quant_kernel(tensor,self.fixed_size,
                                    self.fixed_prec)
    \end{lstlisting}
\end{figure}
The core of the toolbox are the quantizer objects, which carry out the quantization of the tensors. Their simple interface takes a tensor and outputs the quantized version. 
An example is given in listing \ref{lst:quantizer}. The Quantizer interface forces a ''quantize'' method, which invokes the quantization kernel. The quantization process is carried out by the kernel, which is written in C++. It is possible to write the quantizers entirely in Python, although in the case of intrinsic quantization, this utilizes prohibitively many resources.

\paragraph{Changes to the User's Topology File}
Quantizers can be applied to any node of the topology, but it would be very laborious to assign them by hand. For the \Tensorflow \Slim framework, a series of convenience functions is implemented in \TensorQuant, which automate the application of quantizers to the topology. An example of a file describing a topology is given in listing \ref{lst:network_original}.
\begin{figure}
    \begin{lstlisting}[caption={Example topology file (original).},label={lst:network_original}]
def lenet(images, ...):
    ...
    with tf.variable_scope(scope, 'LeNet', 
    [images, num_classes], ...):
    net = slim.conv2d(images, 32, [5, 5], scope='conv1')
    net = slim.max_pool2d(net, [2, 2], 2, scope='pool1')
    net = slim.conv2d(net, 64, [5, 5], scope='conv2')
    ...
    logits = slim.fully_connected(net, num_classes, 
    activation_fn=None, scope='fc4')
    ...
    return logits, end_points
    \end{lstlisting}
\end{figure}
The changes to the topology file when prepared for quantization are shown in listing \ref{lst:network_quant}. 
\begin{figure}
    \begin{lstlisting}[caption={Example topology file (for quantization).}, label={lst:network_quant}]
def lenet(images, ...,
    conv2d=slim.conv2d,  # added layer types
    max_pool2d=slim.max_pool2d, 
    fully_connected = slim.fully_connected):
    ...
    with tf.variable_scope(scope, 'LeNet', 
    [images, num_classes],...):
    # removed slim. before layer calls
    net = conv2d(images, 32, [5, 5], scope='conv1')
    net = max_pool2d(net, [2, 2], 2, scope='pool1')
    net = conv2d(net, 64, [5, 5], scope='conv2')
    ...
    logits = fully_connected(net, num_classes, 
    activation_fn=None, scope='fc4')
    ...
    return logits, end_points
    \end{lstlisting}
\end{figure}

\paragraph{Assigning Layers for Quantization}
\ifx false
\begin{figure}
    \begin{lstlisting}[caption={Example Quantization Dictionary},label={lst:dict}]
    intrinsic_dictionary={
    'conv1':fixed_point_quantizer_object_1,
    'pool1':other_quantizer_object_2
    }
    \end{lstlisting}
\end{figure}
\fi
The locations where quantization is applied are controlled with the \Tensorflow variable namespaces. The user has to specify the entire variable name for a single, or matching substrings for a set of layers where the quantization should be applied to. For example, in listing \ref{lst:network_quant} there is the variable scope ''LeNet'', which contains the layers ''conv1'', ''pool1'' and so on. The first convolution layer can be accessed with the identifier \mbox{''LeNet/conv1''}. All layers in the ''LeNet'' scope can be accessed at once by the identifier ''LeNet''.\\
The user specifies two dictionaries, one for intrinsic and one for extrinsic rounding. The keys are the identifiers, and the values are quantizer objects. 
The dictionaries are passed to the layer factories.

\paragraph{The Layer Factories}
\label{sec:layer_factories}
\begin{figure}
\begin{lstlisting}[caption={Pseudocode for layer factory.}, label={lst:layer_factory}]
def layer_factory(intrinsic_dictionary, 
                    extrinsic_dictionary):
  def func(*args,**kwargs):
    current_layer=get_current_layer()
    if current_layer in intrinsic_dictionary:
        kwargs['quantizer']=
               intrinsic_dictionary[current_layer]
        net=quantized_conv(*args,**kwargs)
    else:
        net=standard_conv(*args,**kwargs)

    if current_layer in extrinsic_dictionary:
        return extrinsic_dictionary[current_layer].
                                        quantize(net)
    else:
        return net   
  return func
\end{lstlisting}
\end{figure}
The layers are built by factory functions. Each layer type has its own factory that returns a function, which has the same interface as the standard \Tensorflow layers. The factory takes two dictionaries as input arguments, one for intrinsic and extrinsic quantization. The pseudocode of a factory function is given in listing \ref{lst:layer_factory}.

\paragraph{Implementation of Extrinsic and Intrinsic Quantization}
\ifx false
\begin{figure}
    \begin{lstlisting}[caption={Pseudocode for convolution reimplementation},label=lst:convolution]
    def convolution_op(input, filter, 
    quantizer):
    ...
    net = multiply(input,filter)
    # quantize after multiply
    net = quantizer.quantize(net) 
    net = reduce_sum(net)
    # quantize after add
    output = quantizer.quantize(net)
    return output
    \end{lstlisting}
\end{figure}
\fi
Implementing extrinsic quantization is straight forward, because the quantization is applied directly to the layer output. The quantization is independent of what is computed in the layer, thus it can be applied directly to all layer types.\\ 
In the intrinsic case, however, there is no such straight forward approach since the specific operation of the layer needs to be considered. Unfortunately, there is no other way than to re-implement the layer type, where quantization is applied to the desired calculation steps.\\
The aim is to use the same quantizer objects as previously shown in listing \ref{lst:quantizer}, thus no additional C++ kernels need to be written. In the re-design, the standard \Tensorflow framework is used as much as possible. 
The downside is, that a lot of intermediate results need to be stored in tensors, which means an increase in required memory. To mitigate this problem, the batch size can be reduced.\\
Adding extrinsic quantization to the model increases the time to build the model by less than 2\,\% and the runtime by 20\,\% (estimated on \InceptionI). Intrinsic quantization increases the build time by a factor of approximately $\times \,70$ and the runtime by $\times \,20$.\\

Some layers contain trained parameters, e.g. filter weights. For a proper representation of fixed point operations, those values are automatically quantized with the intrinsic quantizer before passed to the calculation.
Adding quantization does not interfere with the \Tensorflow namespaces. Therefore, the model parameters can be loaded from save files where the model was trained without quantization.\\
Extrinsic and gradient quantization are independent from the \Tensorflow version. Intrinsic quantization needs to re-implement layers, therefore it can be incompatible to other versions than 1.0 and 1.2.

\paragraph{Gradient Quantization}
\ifx false
\begin{figure}
    \begin{lstlisting}[caption={Quantization of the gradient},label={lst:gradient}]
    ...
    gradients=[(gradient_quantizer.quantize(gv[0]),gv[1])
    for gv in gradients]
    gradient_updates = optimizer.apply_gradients(
    gradients,...)
    ...
    \end{lstlisting}
\end{figure}
\fi
The gradient quantization as described in equation \ref{eq:grad_quantization} is implemented easily. In the file controlling the training, the gradients are intercepted and the quantizer is applied, before they are passed to the optimizer function.\\


\section{Experiments}
\label{sec:experiments}
The \TensorQuant toolbox is used to apply fixed point quantization to DNNs. The simulations are focused on popular CNN topologies, such as \InceptionI (GoogleNet) \cite{SzeLiu15}, \InceptionII \cite{SzeVan16}, \ResnetI and \ResnetII \cite{HeZha16}. The networks are trained on ImageNet 2012 \cite{imagenet_cvpr09, ImageNet2012}. 
For reference, we also provide results for LeNet \cite{LecBot98} with the MNIST dataset. In the learning experiments, AlexNet \cite{KriSut12} is trained on ImageNet.
The networks and trained parameters are taken from the \Tensorflow model library\cite{TensorflowModels}, specifically the \Slim {\it GitHub} web page \cite{Slim}.\\
The main metric is the test accuracy. An unquantized version of each topology serves as the baseline, to which all accuracies are given to as relative values. About 1\% of the validation set is used for inference, simply to perform design space exploration in reasonable time, especially in the case of intrinsic quantization. Using a smaller validation set is valid, because it is only interesting whether or not the baseline accuracy is reached, plus quantization without any form of retraining cannot improve accuracy.

\subsection{Inference with Quantization}
The following experiments perform inference on pre-trained networks. The files with the pre-trained parameters are downloaded from the \Slim web page \cite{Slim}.

\subsubsection{Rounding Method}
\begin{figure} [htbp]
	\centering
	\includegraphics[width=0.9\linewidth]{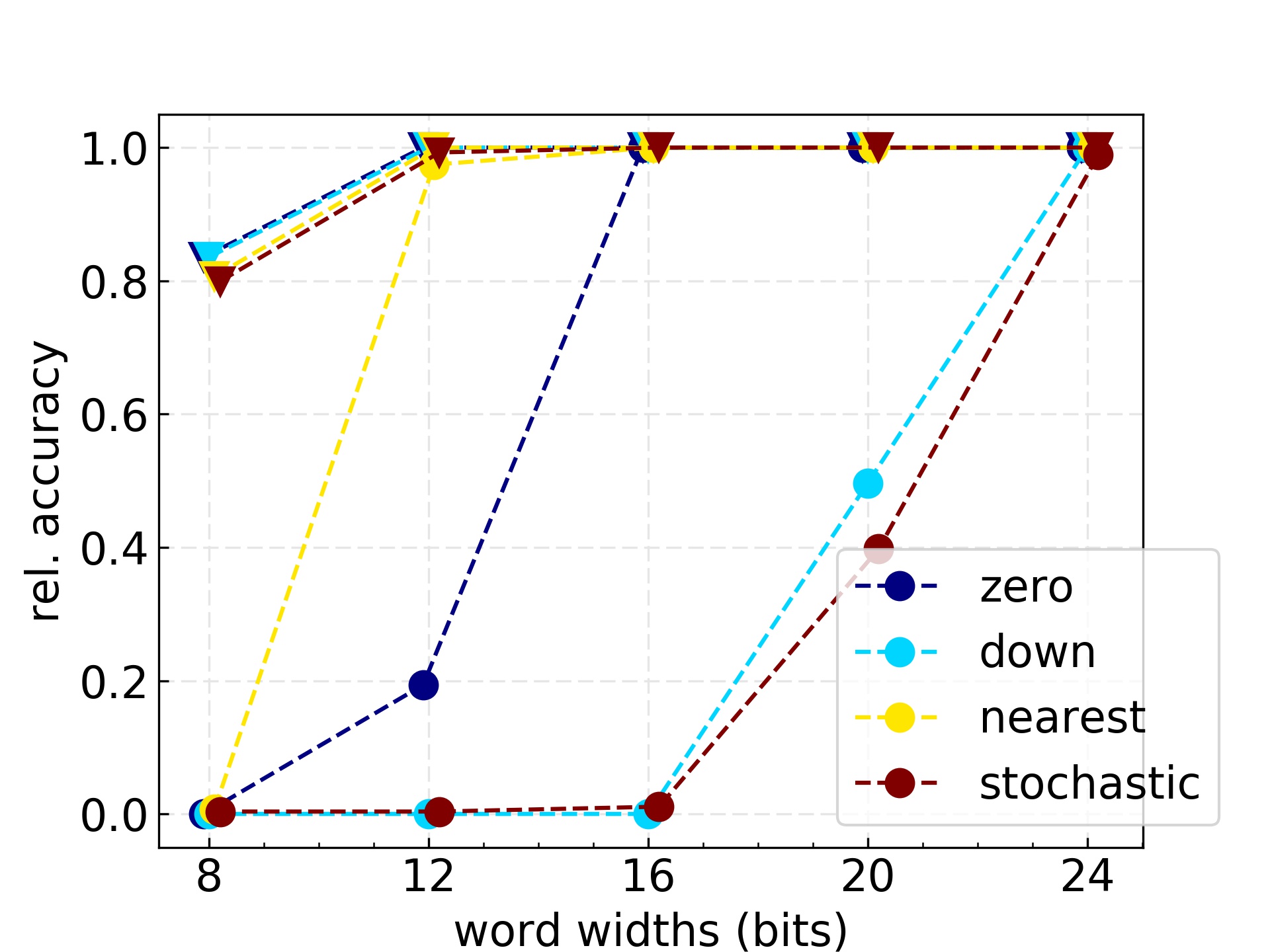}
	\caption{ Zero, down, nearest and stochastic rounding methods applied to the \InceptionI topology. Dots mark the relative accuracy when the rounding is applied intrinsically and triangles mark the extrinsic rounding.}
	\label{fig:inceptionv1_rounding}
\end{figure}

\begin{figure} [htbp]
	\centering
	\includegraphics[width=0.9\linewidth]{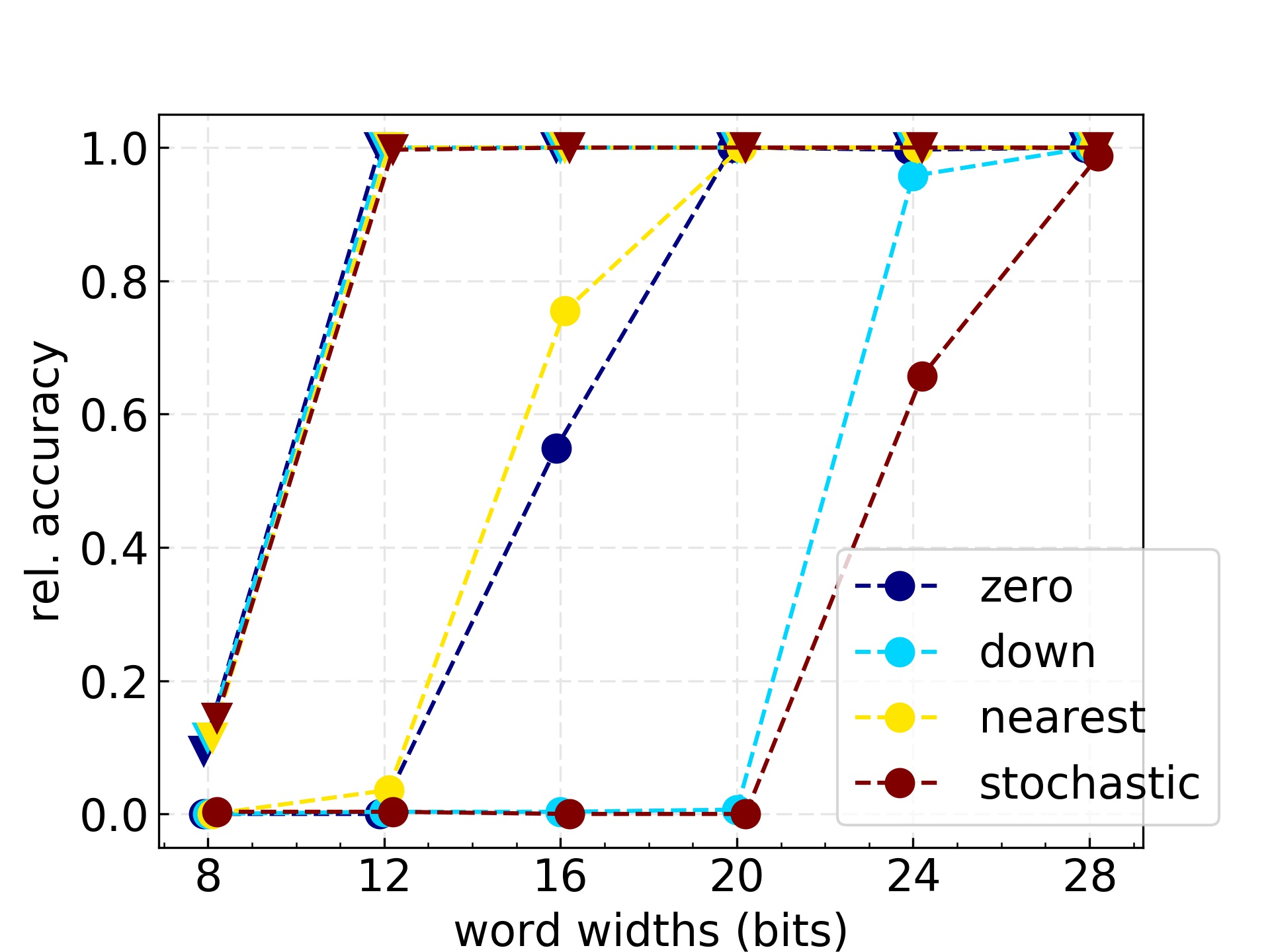}
	\caption{Rounding methods applied to the \ResnetI topology, as in figure \ref{fig:inceptionv1_rounding}.}
	\label{fig:resnet50_rounding}
\end{figure}
The influence of the different rounding methods introduced in section \ref{sec:fixed_point} is investigated on the \InceptionI and \ResnetI topologies. Figures \ref{fig:inceptionv1_rounding} and \ref{fig:resnet50_rounding} show the relative accuracy against the word width used for rounding. The fractional bits are half of the word size. Intrinsic and extrinsic rounding are plotted in the same figure with different markers.\\
Extrinsic rounding is almost not affected by the rounding method. The only deviation from the baseline accuracy is coming from the word size. All rounding methods lead to equally good accuracies.\\
For the intrinsic case, the choice of the rounding method has an impact on the accuracy. Using down or stochastic rounding in the \InceptionI topology doubles the required word size compared to nearest rounding, whereas zero rounding is between those. For the \ResnetI topology, nearest and zero rounding lead to similar accuracies, but down ans stochastic rounding are still worse than the other two methods.\\
Nearest rounding is the method which requires the least amount of bits, therefore it is the best method of the investigated ones. It is used in the following experiments. Down and stochastic rounding, on the other hand, are the most demanding methods.\\

\subsubsection{Intrinsic vs. Extrinsic Rounding}
\begin{table}
	\caption{Comparison of rounding all layers at once (all) against the subunit method (single). The entries are (relative accuracy, fractional bits).}
	\label{tab:layerVsAll}
	\begin{tabular}{lcccc}
	\toprule
	 Topology & 8\,bit & 12\,bit & 16\,bit & 20\,bit \\ 
	\midrule
			\InceptionI (all)  &0.01, 4 &  0.99, 6 &  1.00, 8 &  1.00, 8 \\ 
	    	\InceptionI (single)  &  0.46, 2 &   1.00, 8 &   1.00, 8 &  1.00, 8\\ 
	    	\\
			\ResnetI (all)  &0.00, 0 &  0.04, 6 &  0.91, 6 &  1.00, 10 \\ 
			\ResnetI (single)  &  0.02, 2 &  0.54, 4 &   0.99, 6 &  1.00, 10\\ 
	\bottomrule
	\end{tabular}
\end{table}

For intrinsic rounding, the subunit approach explained in section \ref{sec:subunit_quantization} is used. Table \ref{tab:layerVsAll} compares the accuracies from rounding all layers at once against the subunit approach for \InceptionI and \ResnetI.
If the relative accuracy is very close to 100\,\% in the subunit approach, then the method is equivalent to rounding the entire network at once, as the results are the same.

\begin{table}
	\caption{ Comparison of different topologies and word sizes for intrinsic rounding. The entries are (relative accuracy, fractional bits).}
	\label{tab:intrinsic_all}
	\begin{tabular}{lccccc}
	\toprule
	Topology &8 bit &10 bit &12 bit &14 bit &16 bit \\ 
	\midrule
        \InceptionI  &  0.46, 2 &  0.98, 5 &  1.00, 8 &  1.00, 9 &  1.00, 8\\ 
    	\InceptionII &  0.08, 2 &  0.92, 7 &  0.97, 6 &  0.99, 7 &  0.98, 6\\ 
		\ResnetI  &  0.02, 2 &  0.14, 3 &  0.54, 4 &  0.93, 5 &  0.99, 6\\ 
		\ResnetII  &  0.03, 2 &  0.16, 3 &  0.55, 4 &  0.95, 5 &  1.00, 6\\ 
	\bottomrule
	\end{tabular}
\end{table}
 
\noindent\textbf{Intrinsic} rounding is applied to DNN topologies in table \ref{tab:intrinsic_all}. The word size is fixed in each column. The entries show the maximum achievable relative accuracy and the used fractional bits.\\
Inception type topologies require less word size than ResNet type ones to achieve full benchmark accuracy. 12\,bits are enough for Inception, whereas ResNets need 16\,bits. This is a hint that the different topology types have different bottlenecks, even though they use the same layer types. The bottlenecks will be located later in section \ref{sec:layerwise_intrinsic}.\\
The conclusion from comparing the Inception and ResNet type topologies amongst themselves is that the word size does not depend on the depth of the topology. This can be attributed to the batch normalization layers, which normalize the activations before they leave a layer. The range of the activation values is kept the same, therefore fixed point quantization is working well.\\
\begin{table}
	\caption{Comparison of different topologies and word sizes for extrinsic rounding, similar to table \ref{tab:intrinsic_all}.}
	\label{tab:extrinsic_all}
	\begin{tabular}{lccccc}
	\toprule
	Topology &4 bit &6 bit &8 bit &10 bit &12 bit \\ 
	\midrule
		\InceptionI  &  0.32, 0 &  0.76, 2 &  0.98, 2 &  1.00, 4 &  1.00, 4\\ 
		\InceptionII  &  0.08, 0 &  0.93, 2 &  0.97, 2 &  0.99, 4 &  1.00, 6\\ 
		\ResnetI  &  0.06, 0 &  0.62, 0 &  0.99, 2 &  1.00, 4 &  1.00, 4\\ 
		\ResnetII  &  0.07, 0 &  0.55, 0 &  1.00, 2 &  1.00, 2 &  1.00, 2\\ 
	\bottomrule
	\end{tabular}
\end{table}

\noindent The results from rounding the topologies \textbf{extrinsically} are shown in table \ref{tab:extrinsic_all}, which is structured as in the intrinsic case. The subunit method is not needed here, because extrinsic rounding does not require as much memory to run, therefore all layers can be quantized at once without problems. Notice that the word sizes in the columns are different.
Extrinsic rounding achieves baseline accuracy with less word size than in the intrinsic case. 8 to 10\,bits are already enough to stay close to full accuracy, regardless of the topology type. Also, the portion of fractional bits is lower than in the intrinsic case, meaning the output values of the layers can be transferred in low precision.\\
\begin{table}
	\caption{LeNet, \InceptionI and \ResnetI rounded intrinsically for different word sizes. The entries are (word size, fractional bits).}
	\label{tab:LeNet_intrinsic}
	\begin{tabular}{lcccccccc}
	\toprule
	Topology &4 bit &8 bit &12 bit &16 bit &20 bit\\ 
	\midrule
		LeNet  &  0.40, 3 &  0.90, 4 &  1.00, 6 &  1.00, 8 &  1.00, 8 \\ 
		\InceptionI  &- &  0.01, 4 &  0.99, 6 &  1.00, 8 &  1.00, 8 \\ 
		\ResnetI  &- &  0.00, 0 &  0.04, 6 &  0.91, 6 &  1.00, 10 \\ 
	\bottomrule
	\end{tabular}
\end{table}

\noindent Last, LeNet is compared to the \InceptionI and \ResnetI topologies. All networks are quantized completely, so the subunit method is not used. In table \ref{tab:LeNet_intrinsic}, the results are presented as in table \ref{tab:intrinsic_all} before. However, LeNet was trained on MNIST and the other topologies on ImageNet.\\
At 8\,bits word width, LeNet achieves 90\,\% relative accuracy, but the other topologies are close to zero. As from the previous results, \ResnetI needs more word width than \InceptionI.
This illustrates that even though all three topologies are CNNs and utilize the same layer types, results cannot be generalized from one topology to another for the intrinsic case. 
\subsubsection{Layerwise Intrinsic Rounding}
\label{sec:layerwise_intrinsic}
After seeing the results from the subunit approach in table \ref{tab:intrinsic_all}, the actual per subunit requirements regarding word size and fractional bits are shown in the tables \ref{tab:inception_layerwise} and \ref{tab:resnet_layerwise} for the \InceptionII and \ResnetI topology, respectively. Each column represents a threshold for the relative accuracy. The entries state the minimum required word size and fractional bits to be above the threshold.\\
\begin{table} [ht]
    \centering
	\caption{Subunits and relative accuracies for the \InceptionII topology. The entries show the required  word size and fractional bits to achieve the relative accuracy of the column.}
	\label{tab:inception_layerwise}
	\begin{tabular}{lccc}
	\toprule
	Subunit & 100\% & 99\% & 90\% \\ 
	\midrule
		Conv2d\_1a\_3x3  & 24,6 & 16,12 & 12,6\\ 
		Conv2d\_2a\_3x3  & 16,6 & 12,6 & 8,2\\ 
		Conv2d\_2b\_3x3  & 12,6 & 12,6 & 8,4\\ 
		MaxPool\_3a\_3x3  & 8,4 & 8,4 & 8,0\\ 
		Conv2d\_3b\_1x1  & 12,6 & 12,6 & 8,4\\ 
		Conv2d\_4a\_3x3  & 8,4 & 8,4 & 8,4\\ 
		MaxPool\_5a\_3x3  & 8,0 & 8,0 & 8,0\\ 
		Mixed\_5b  & 8,4 & 8,4 & 8,4\\ 
		Mixed\_5c  & 12,4 & 12,4 & 8,4\\ 
		Mixed\_5d  & 16,10 & 16,10 & 8,4\\ 
		Mixed\_6a  & 8,4 & 8,4 & 8,4\\ 
		Mixed\_6b  & 12,6 & 12,6 & 8,4\\ 
		Mixed\_6c  & 12,6 & 12,6 & 8,6\\ 
		Mixed\_6d  & 12,8 & 12,8 & 12,6\\ 
		Mixed\_6e  & 12,8 & 12,8 & 12,6\\ 
		Mixed\_7a  & 24,20 & 16,8 & 8,4\\ 
		Mixed\_7b  & 8,6 & 8,6 & 8,6\\ 
		Mixed\_7c  & 8,6 & 8,6 & 8,6\\ 
		AuxLogits  & 8,0 & 8,0 & 8,0\\ 
		PreLogits  & 8,0 & 8,0 & 8,0\\ 
		Logits  & 12,4 & 12,4 & 12,4\\ 
	\bottomrule
	\end{tabular}
\end{table}

\begin{table} [htb]
    \centering
	\caption{Subunits and relative accuracies for the \ResnetI topology, as in table \ref{tab:inception_layerwise}.}
	\label{tab:resnet_layerwise}
	\begin{tabular}{lccc}
	\toprule
	Subunit & 100\% & 99\% & 90\% \\ 
	\midrule
		resnet\_v1\_50/conv1 & 20,8& 16,6& 16,6\\ 
		block1 & 16,10& 12,6& 12,6\\ 
		block2 & 16,10& 12,6& 12,6\\ 
		block3 & 12,8& 12,8& 12,6\\ 
		block4 & 16,10& 16,8& 12,6\\ 
		logits & 16,8& 12,4& 12,4\\ 
	\bottomrule
	\end{tabular}
\end{table}

From this perspective, one can identify the bottleneck units and how demanding they are. Most units in both Inception and ResNet topologies have the same requirement of 12\,bits word width and 6\,bits fractional part, because most of the layers within a subunit are convolution layers. 
However, the bottleneck layers of the network determine the required word width of the entire hardware architecture. The very first layer needs a high word size in both topologies. In \InceptionII, there are bottlenecks appearing in the middle of table \ref{tab:inception_layerwise}. There is no general rule that bottlenecks appear at the beginning or the end of a topology.\\

\ifx false
\subsubsection{Batch Normalization}
The batch normalization has been excluded from the previous experiments. However, it is a vital element of the networks which use it and should not be neglected. The minimal settings which lead to a relative accuracy of above 99\,\% for intrinsic rounding are shown in table \ref{tab:crosseval_batch}. 
\begin{table}
	\caption{ The needed w,f for batch normalization(rel. accuracy deviates max. 0.01)}
	\label{tab:crosseval_batch}
	\begin{tabular}{lc}
	\toprule
	Net & w,f \\ 
	\midrule
		inception\_v1  & 12,6\\ 
		inception\_v3  & 12,8\\ 
		resnet\_v1\_50  & 28,10\\ 
		resnet\_v1\_152  & 28,10\\ 
	\bottomrule
	\end{tabular}
\end{table}

It may be surprising that the batch norm layers of the Resnet type networks need so much more word size. The reason for that is simply that the learned variable for the variance in the batch norm layers tends to be in the thousands for Resnet type, whereas it is between zero and one for Inception type networks. Resnets therefore need higher word widths to contain those values and the intermediate results in the computation of the batch norm layer. This may present a problem for hardware implementation.\\
\fi

\subsection{Training with Quantization}
For the case of distributed systems, it is more interesting to see if and how well a topology can be trained when the gradients, which are communicated between the computation nodes, are quantized. As before, the quantization method is fixed point with nearest rounding. The accuracies are relative to a network trained with exactly the same set up, but without gradient quantization.\\
Table \ref{tab:LeNet_train_accuracy} shows the relative test accuracies for different word size and fractional bits combinations for LeNet, and table \ref{tab:Alexnet_train_accuracy} for AlexNet.
\begin{table}
	\caption{Relative accuracy achieved for LeNet when the gradient is quantized with different word sizes (W) and fractional bits (F).}
	\label{tab:LeNet_train_accuracy}
	\begin{tabular}{lcccccc}
	\toprule
	(W,F) &2,1 &4,2 &6,3 &8,4 &10,5 &12,6 \\ 
	\midrule
		rel. accuracy  &  0.87 &  0.97&  0.99 &  0.99 &  1.00 &  1.00 \\ 
	\bottomrule
	\end{tabular}
\end{table}

\begin{figure}
    \centering
    \includegraphics[width=0.9\linewidth]{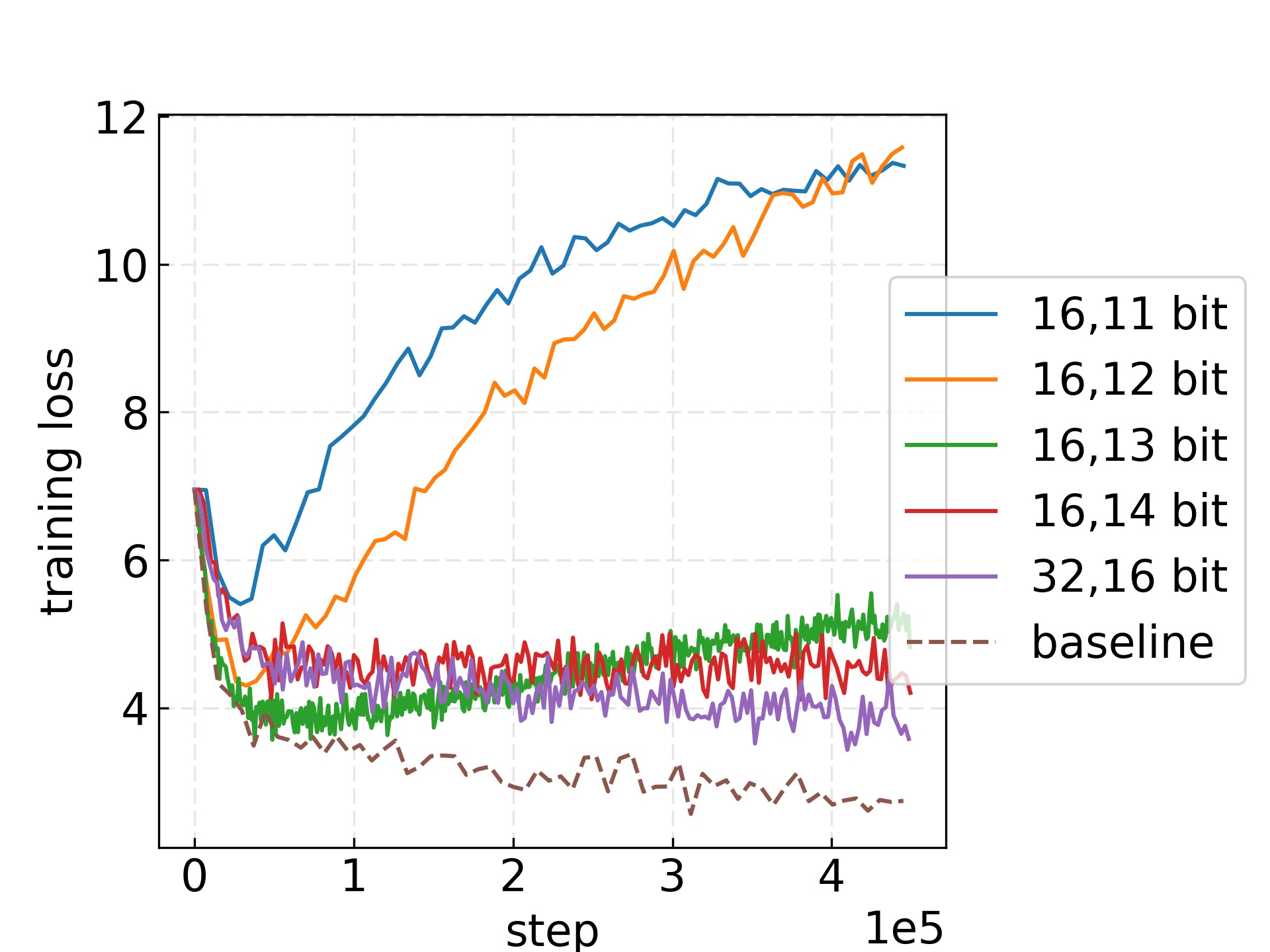}
    \caption{Training loss versus number of steps for gradient quantization and the AlexNet topology.}
    \label{fig:train_loss}
\end{figure}

\begin{table}
    \caption{Relative accuracy achieved for AlexNet when the gradient is quantized with different word sizes, similar to table \ref{tab:LeNet_train_accuracy}.}
    \label{tab:Alexnet_train_accuracy}
    \begin{tabular}{lccccccc}
        \toprule
        (W,F)  & 16,8  & 16,11 & 16,12 & 16,13 & 16,14 & 32,16 \\ 
        \midrule
        rel. accuracy & 0.004 & 0.847 & 0.926 & 0.997 & 0.745 & 1.0 \\ 
        \bottomrule
    \end{tabular}
\end{table}
The gradient can be quantized quite rigorously in LeNet. Only 4\,bits are sufficient to train LeNet to 97\,\% relative accuracy on the MNIST dataset. AlexNet, on the other hand, needs at least 12 fractional bits to be trained to 92\,\%. The comparison between the two topologies shows that there is no point in generalizing results 
from simple topoligies like LeNet to larger and deeper networks.\\
An interesting observation can be made when looking at the training loss for AlexNet. Figure \ref{fig:train_loss} shows the training loss against the number of steps for the AlexNet topology, trained with the word sizes of table \ref{tab:Alexnet_train_accuracy}.
The training loss is going up after an initial phase of descend for 11 and 12 fractional bits. Despite the training loss going up, the test accuracy is high. The reason for that is that the used function for the training loss comprises of the cross entropy and the L2-regularizer. 
The quantization renders the L2-regularizer ineffective during the computation of the gradient in case of low fractional bits, so only the cross entropy is minimized. Since the weights are not bound by the regularizer, their magnitudes can grow freely. However, then the regularizer contributes a high value to the overall training loss.

\section{Discussion}

A paper which has investigated the fixed point data format with similar thoroughness is \cite{JudAlb15}. The most complex network they investigated is \InceptionI. They suggest using at most 14\,bits word length with 2\,bits fractional part. This result somewhat coincides with our findings for extrinsic rounding, where 10\,bits word size and 4 fractional bits are sufficient for that topology. The difference in the results could come from the used framework and model parameters.\\
Designers of custom hardware have been using 16\,bit word size for implementations of their topologies \cite{HanLiu16, GokZai17, ChenKri17}.
Our results suggest to be careful, because there is no general word size which guarantees to be sufficient for all topologies. But for the investigated cases, 16\,bit is large enough for all layers.\\
The results from the training coincide with \cite{GupAgr15}, although they used extrinsic rounding, instead. The fractional part needs to be relatively high, whereas the integer part is not needed. 
The increase of the training loss when using 16\,bit nearest rounding during training (figure \ref{fig:train_loss}) was also observed by \cite{GupAgr15}, despite using extrinsic quantization. In their paper, stochastic rounding makes the training loss converge normally.

\section{Conclusion}
The \TensorQuant toolbox allows to explore different quantization methods with the \Tensorflow framework. The unique feature is that the fixed point data format can be emulated to the arithmetic level, thus allowing for the closest similarity to custom hardware yet presented in any framework.\\
The most important result from the experiments is that in the case of intrinsic rounding, the word size is more demanding than it was previously thought. Intrinsic rounding is very sensitive to the rounding method. It is not possible to generalize the required word width from one topology to another, as they have different bottleneck layers. The results from the training section show that gradient quantization allows even less to generalize between topologies.\\
It is planned to extend the \TensorQuant toolbox much further, especially the functionality related to training. The goal is to fully emulate learning on a distributed system comprising of custom hardware, thus using the fixed point data format. 
Other quantization methods will be implemented to study possible strategies to reduce bandwidth in distributed systems. 
Also, \TensorQuant will be applied to other topologies like recurrent neural networks. It is expected that the requirements regarding the data representation will differ from CNNs.

\section*{Acknowledgments}
\noindent The authors thank the Leibniz Supercomputing Centre for providing compute time on their DGX-1 system. 
We also gratefully acknowledge the support of the NVIDIA Corporation
with the donation of the Titan X Pascal GPU used for this research.

\bibliographystyle{acm}
\bibliography{MAIN}

\end{document}